# A Review on Domain Adaption and Generative Adversarial Networks(GANs)

**Aashish Dhawan**
UBTECH AI Center
University of Sydney, Sydney
NSW, Australia
aashudhawan@gmail.com

**Divyanshu Mudgal**
Computer Science Engineering
JMIETI, Radaur
Yamunanagar, India
div18mudgal@gmail.com

**Vishal Garg**
Computer Science Engineering
JMIETI, Radaur
Yamunanagar, India
vishalgarg@jmieti.org

*Abstract*— the major challenge in today's computer vision scenario is the availability of good quality labeled data. In a field of study like image classification, where data is of utmost importance, we need to find more reliable methods which can overcome the scarcity of data to produce results comparable to previous benchmark results. In most cases, obtaining labeled data is very difficult because of high cost of human labor and in some cases impossible. The purpose of this paper is to discuss about Domain Adaption and various methods to implement it. The main idea is to use a model trained on a particular dataset to predict on data from a different domain of the same kind, example – model trained on paintings of airplanes predicting on real images of airplanes.

*Keywords—Domain Adaption; Computer Vision; Image classification*

## I. INTRODUCTION

Image Classification refers to the process in computer vision that can classify an image based on its visual content, for example an image classification algorithm may be designed to tell if an image contains a Vehicle or not. It looks like a simple task for a human, but it is quite an arduous task for a computer to recognise the visual content of images. Today most big enterprises in the world are using Image classification to enhance the user experience, like Facebook identifying the people in an image and notifying them about it or mobile phones classifying different faces in an image. Image Classification can be defined as the process of taking an input (image) and outputting a class (like a "human" or a "cat").

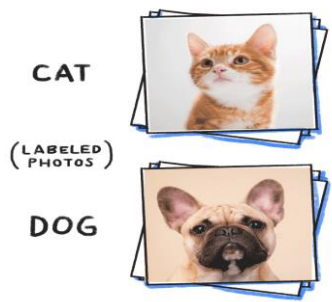

**Figure 1. Image Classification Example**

The process of image classification involves defining a set of classes (objects in the given set of images) and training a model to identify them using labelled dataset. Convolutional Neural Networks (CNNs) are used mostly for the image classification tasks in real world problems. Main reason being that CNNs give very precise results if trained on high-quality annotated data. In a regular classification problem, you would often use one of the standard networks (ResNET, VGG etc.) and build your network on it to train on the specific dataset. The accuracy of a CNN depends upon the amount of data available for it to be trained. The other major drawback with this technique in real world is the problem of domain shift.

The model assumes that the training and testing data both are in the same domain and come from the same underlying distribution. But when the domain shifts, for example a autonomous driver trained on the streets of New York will become a failure if tested in Paris. You have a model trained on the dataset from NYC. It knows what lies ahead (ex – trees, cars, pedestrian etc). Testing the model on streets of Paris will be a disaster and anything and everything could go wrong. The Autonomous driver won't be able to detect traffic lights anymore. The model won't be able to recognize buses and taxis as they look bit different in both the cities.

The reason the model did not work well in this problem is that the problem domain changed. Similarly, you could try to train a model on a dataset of sketches and test it on real life images. This is known as Domain Shift. In these type of scenarios, Domain Adaption comes to rescue. Domain Adaption uses labelled data in one or more domains to predict in the target domain.

In this paper, we will discuss about Domain Adaption and a few different approaches Domain Adaption has been used in the past two years with results from various experiments and what more can be done to enhance the results obtained from Domain Adaption. We will discuss about different types of Generative Adversarial Networks (GANs) in detail.

## II. DOMAIN ADAPTION

By Domain Adaption, we mean an algorithm that is trying to transfer between two domains, usually known as source and target (for instance paintings and real photos). To do so, we can either translate one domain into another or find a common embedding between the domains. In real world problems, we have labelled data from one dataset and need to predict on a different target dataset. This is called unsupervised domain adaption, and this is where domain adaption has been able to record a large amount of success. Besides image classification, domain adaption also has a lot of applications in NLP too.

The most common problem in domain adaption is to predict output of SVHN (dataset made with house numbers) by using MNST (handwritten number dataset).

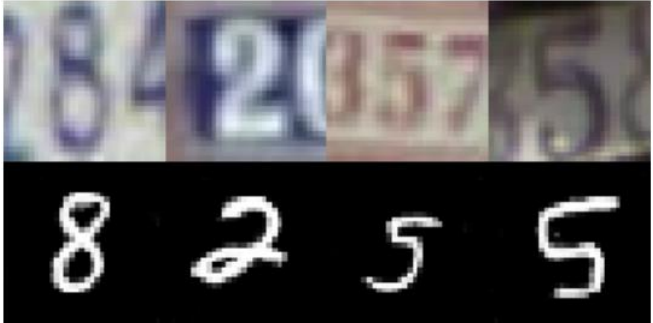

**Figure 2. Transfer of SVHN to MNST**

The results on this problem have improved from 82% accuracy using DRCN[1] to 99.2% with self-ensembling domain adaption[2].

### A. Adversarial Domain Adaption

The main idea behind Adversarial domain adaption (ADA) is using two different neural networks: a Discriminator; that distinguishes between the target domain and the source domain and a Generator that tries to fool the discriminator to make the source and target domains look identical [3]. Both the Generator and Discriminator are neural networks Generator generates sample dataset and feeds it directly to the discriminator as input.

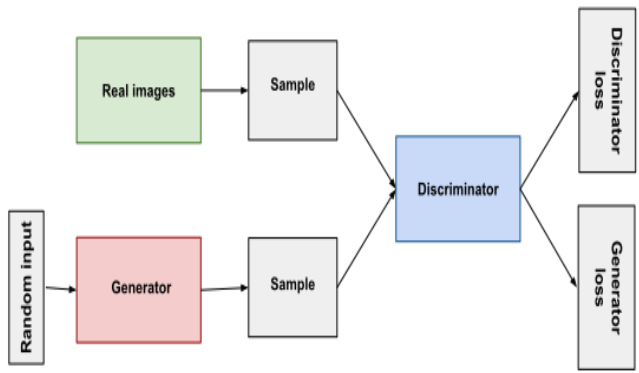

**Figure 3. Basic Structure of GANs**

Through backpropagation, discriminator provides signals used to updates its weights.

The discriminator gets its input from two sources: real data from the actual dataset and the fake data created by the generator. The real data is used as positive instance and the fake data is used as negative instance during the training process. The discriminator classifies both the real and fake data from the generator and discriminator loss penalizes it for misclassifying a real as fake or a fake as real. The weights are updated through the backpropagation from the discriminator network.

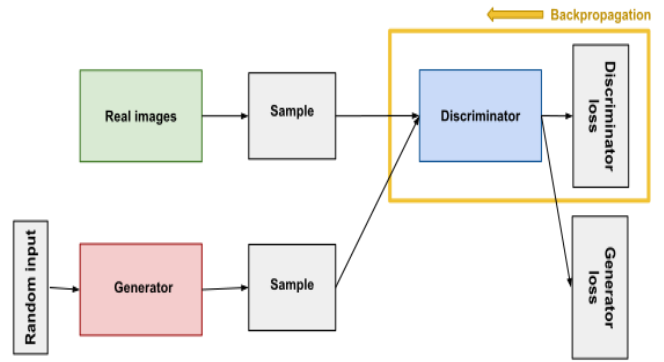

**Figure 4. Backpropagation in Discriminator**

The generator is the part of network that creates fake data by taking feedback from the discriminator. The main aim for a generator is to learn to make the discriminator classify its output as real. The generator is also a neural network and neural networks require some input in order to work. So, a random input is given in its first step and generator turns it into a data instance to be passed to the discriminator as input. After this, the discriminator network classifies the generated data as real or fake and the generator loss penalizes the generator for failing to fool the discriminator.

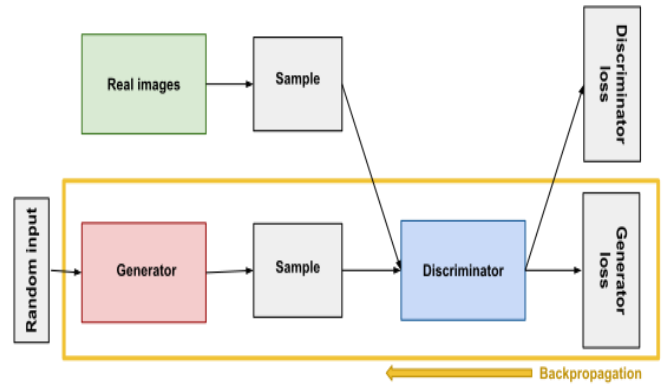

**Figure 5. Backpropagation in Generator**

The training process continues until the generator is able to generate data samples as close to the real data and the discriminator is no more able to differentiate between them.

There are many different ways to implement domain adaption, example – DANN [4], CycleGAN[5],

DiscoGAN[6] etc. We will discuss some of the ideas in detail further.

A. CycleGAN

The main idea of CycleGAN is to train two conditional GANs, one that transfers source to target, and the other target to source. It considers a new kind of loss called cycle-consistency. By connecting the two models together, it tries to produce a mapping between the source and target (source-target-source). They used it to convert horses to zebras and winter scene to summer scenes and produce some amazing results.

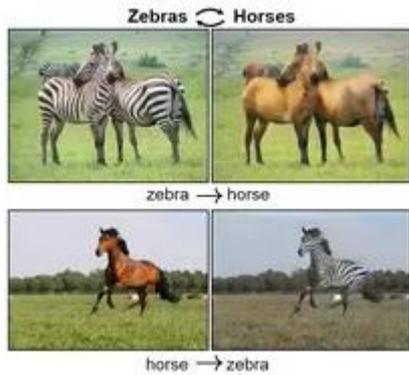

Figure 6. Horse to Zebra transfer using CycleGAN

They trained their model on two different set of distributions instead of picture to picture (sketch and photo of the same object) transfer to obtain a more generalized model, which makes the result even more impressive.

B. Self-Ensembling of Domain Adaption

Apart from adversarial domain adaption, few other methods have been tested. This method tries to find a common embedding between the source and target, forming a single neural network that can classify both source and target domains. Self ensembling method [2] surpassed all the benchmarks on SVHN-MNIST dataset with a 99.2% accuracy. It exploits the mean-teacher method [7].

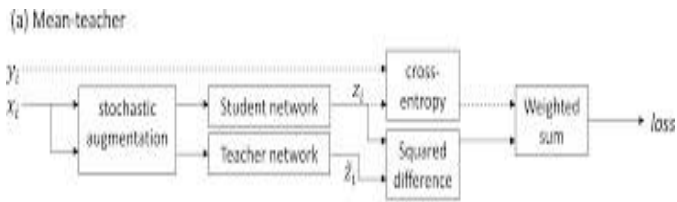

Figure 7. Mean Teacher method

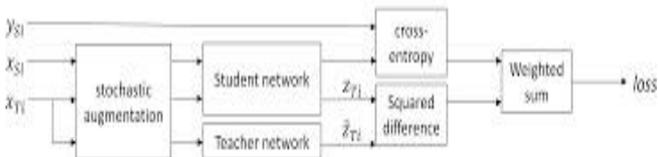

Figure 8. Self-Ensembling Architecture

This model also won the VisDA 2017 challenge and performed incredibly well on small image benchmarks.

C. Domain-Adverdarial Neural Network (DANN)

This is the most interesting model we found in our study. It tried to get rid of the generator and perform domain adaption in a single go. It made use of domain-confusion loss. This is similar to the discriminator loss in GANs. It tries to "confuse" the classifier into thinking that both the domains are same in the high-level classification layer. It contains a gradient reversal layer that matches the feature distribution between the domains. This makes the samples indistinguishable for the classifier.

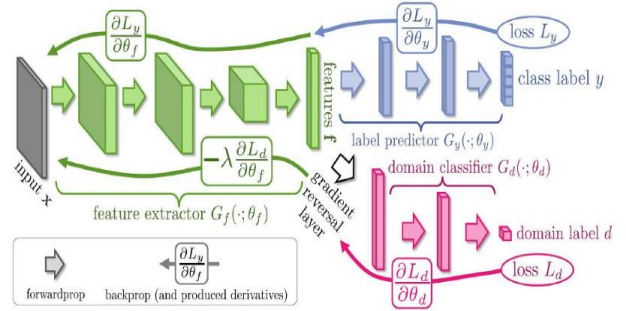

Figure 9. DANN Architecture

It contains two main types of losses: Domain confusion loss and the classification loss. Main idea is to minimize the classification loss for the source samples and the domain confusion loss for all samples (while maximizing the domain confusion loss for the feature extraction). This makes the classifier think of both domains as same and results in a higher accuracy.

III. CONCLUSION

After all the research and study, it is clear that domain adaption has grown to be very impressive in the image classification task. But most of the other applications of domain adaption and GANs have not been yet utilized or discovered. There can be a lot of work done in the field of NLP with the help of domain adaption. For example – Word Translation using DA – translating words in one language (source) to another (target) without the help of labels.

Other important step to be taken using domain adaption is the multi-source domain adaption. As the example given in the paper of autonomous car, we need a car that can drive in all the countries not just two and every country has different set of traffic signs and different roads.

IV. REFERENCES

[1] J. Kim, J.K. Lee (2015). "Deep-Recursive Convolutional Neural Networks for Image Super-Segmentation", Conference on Computer Vision and Pattern Recognition. DOI:10.1109/CVPR.2016.181.
[2] G. French, Michal M., Mark Fisher (2017). "Self-Ensembling for Visual Domain Adaption", proceedings of ICLR *2017*.
[3] Ian Goodfellow et al. "Generative Adversarial Networks" (2014). Advances in Neural Information Processing Systems (NIPS 2014).
[4] Y. Ganin, E. Ustinove et al. "Domain Adversarial training of Neural Network". Journal of Machine Learning Research(2016).

[5] Jun Yan Zhu, Taesung Park, Phillip Isola, Alexei A.(2017) "Unpaired Image-to-Image Translation using Cycle Consistent Adversarial Networks", Proceedings of International Conference on Computer Vision (ICCV 2017). Doi : 10.1109/ICCV.2017.244.

[6] T. Kim, M. Cha, H. Kim et al. (2017). "Learning to Discover Cross-Domain Relations with Generative Adversarial Networks". ICML, Proceedings of 34th International Conference on Machine Learning.

[7] A. Tarvainen, Harri V. (2017). "Mean Teachers are better role models: Weight-Averaged Consistency Targets improve semi-supervised deep learning results". Proceesings of NIPS 2017 .